\documentclass{article}
\usepackage{cite}
\usepackage{amsmath,epsfig,amssymb,amsfonts}
\usepackage{algorithmic}
\usepackage{graphicx}
\usepackage{textcomp}
\usepackage{xcolor}
\usepackage{algorithm} 
\usepackage{algorithmic}
\usepackage{multirow}
\usepackage{subfigure}
\usepackage{caption}
\usepackage{soul}
\usepackage{enumitem}

\usepackage[preprint]{spconfa4}

\let\OLDthebibliography\thebibliography
\renewcommand\thebibliography[1]{
  \OLDthebibliography{#1}
  \setlength{\parskip}{0pt}
  \setlength{\itemsep}{0pt plus 0.3ex}
}

\begin{document}\sloppy

\def\x{{\mathbf x}}
\def\L{{\cal L}}

\title{Fairness-Aware Client Selection for Federated Learning}
%
\name{Yuxin Shi$^{1,2,3}$, Zelei Liu$^{1}$, Zhuan Shi$^{1,4}$, Han Yu$^{1}$}
\address{$^{1}$School of Computer Science and Engineering, Nanyang Technological University (NTU), Singapore \\
$^{2}$Alibaba-NTU Singapore Joint Research Institute, NTU, Singapore.
$^{3}$Alibaba Group, Hangzhou, China \\
$^{4}$School of Computer Science and Technology, University of Science and Technology of China, China \\
\{shiy0029, zelei.liu, han.yu\}@ntu.edu.sg, zhuanshi@mail.ustc.edu.cn}

\maketitle

\begin{abstract}
Federated learning (FL) has enabled multiple data owners (a.k.a. FL clients) to train machine learning models collaboratively without revealing private data. Since the FL server can only engage a limited number of clients in each training round, FL client selection has become an important research problem. Existing approaches generally focus on either enhancing FL model performance or enhancing the fair treatment of FL clients. The problem of balancing performance and fairness considerations when selecting FL clients remains open. To address this problem, we propose the \underline{Fair}ness-aware \underline{Fed}erated \underline{C}lient \underline{S}election (FairFedCS) approach. Based on Lyapunov optimization, it dynamically adjusts FL clients' selection probabilities by jointly considering their reputations, times of participation in FL tasks and contributions to the resulting model performance. By not using threshold-based reputation filtering, it provides FL clients with opportunities to redeem their reputations after a perceived poor performance, thereby further enhancing fair client treatment. Extensive experiments based on real-world multimedia datasets show that FairFedCS achieves 19.6\% higher fairness and 0.73\% higher test accuracy on average than the best-performing state-of-the-art approach.

\end{abstract}
\vspace*{-2mm}
\section{Introduction}
\label{sec:intro}
The success of multimedia applications (e.g., computer vision-based systems) depends on the availability of sufficient data to train effective machine learning (ML) models~\cite{chen2022noise}. However, direct access to distributively generated datasets (e.g., usage data from mobile devices) risks infringing on user privacy, leading to the introduction of laws such as the General Data Protection Regulation (GDPR)~\cite{GDPR2017} protection prohibiting such data sharing. Federated learning (FL)~\cite{kairouz2021advances,FL:2019} has been proposed as an alternative collaborative ML paradigm without requiring direct access to sensitive local data in order to adapt to the privacy-aware environment.

With FL come new challenges to be resolved.
One important challenge is data and resource heterogeneity \cite{li2020federated}. During federated model training, data owners (a.k.a. FL clients) contribute not only their local data, but also their computation and communication resources \cite{Nishio2019ClientSF}. As data quality, quantity and local resources vary among clients, their contributions (e.g., impact on model performance, convergence rate) to FL differ. Hence, how to select FL clients to achieve the desired design goals has become an important research problem in the field.

One category of FL client selection approaches focuses on fulfilling the goals of the FL servers. They generally adopt threshold-based approaches to filter out FL clients deemed to be undesirable (e.g., with perceived low-quality data or low reputation)~\cite{Nishio2019ClientSF,luping2019cmfl,tang2022fedcor, cho2022towards, Yoshida2020HybridFLFW}. However, such an approach disregards the FL clients' interest (e.g., opportunities to participate in FL) and may cause some of them to perceive that they are not fairly treated by the FL system. Without the ability to fairly treat FL clients, an FL system risks client attrition which negatively affects the long-term sustainable operation of such systems~\cite{shi2021survey}.
Fairness-aware FL client selection approaches started to emerge in recent years. Some focus on ensuring that data owners have guaranteed rates to be selected to join FL~\cite{Huang2021AnEC}. Others attempt to give more preference to high-performing FL clients while offering equitable opportunities for others to be selected to join FL~\cite{SongReputation22}. Nevertheless, the problem of balancing performance and fairness considerations when selecting FL clients remains open.

In this paper, we bridge this important gap by proposing the \underline{Fair}ness-aware \underline{Fed}erated \underline{C}lient \underline{S}election (FairFedCS) approach. Based on Lyapunov optimization \cite{Yu-et-al:2016,Yu-et-al:2017}, it dynamically adjusts FL clients' selection probabilities by jointly considering their reputations, times of participation in FL tasks and contributions to the resulting model performance. By not using threshold-based reputation filtering, it provides FL clients with opportunities to redeem their reputations even after perceiving poor performance, thereby further enhancing fair client treatment. In addition, it offers an easy-to-use control variable to enable FL system administrators to adjust the trade-off between improving the FL models' performance vs. improving fair treatment of the FL clients.

We conduct extensive experiments based on real-world multimedia datasets (including Fashion-MNIST and CIFAR-10) to evaluate the performance of FairFedCS against four state-of-the-art FL client selection approaches. The results show that FairFedCS achieves 19.6\% higher fairness and 0.73\% higher test accuracy on average than the best-performing state-of-the-art approach.

\vspace*{-1mm}
\section{Related work}
FL client selection approaches can be divided into two main categories: 1) performance-oriented approaches, and 2) fairness-oriented approaches.

Performance-oriented approaches focus on selecting clients that best fulfill the FL servers' goals.
In~\cite{Nishio2019ClientSF,luping2019cmfl}, threshold-based approaches are used to select high-quality FL clients and filter out low-quality ones. The quality metrics used include the amount of local data, contributions to FL model performance and reputation, etc.
However, this brings another problem - oversampling of FL clients from specific groups causing the global FL model to bias towards the data owned by these clients, thereby deteriorating its ability to generalize~\cite{Cho2020ClientSI}. 
Moreover, these approaches do not serve the interests of the FL clients. For an FL system, attracting enough clients to continue participating in FL model training is key to maintaining its long-term sustainable operation. Under the performance-oriented approaches, clients perceived to be of relatively low quality might not be given the opportunity to join FL (a.k.a. unfair selection among clients). As a result, these clients cannot receive any incentives~\cite{incentive_zhan} and might choose to leave the system, resulting in attrition of the pool of candidate FL clients.

Fairness-oriented approaches focus on ensuring that FL clients have equal opportunities to be selected for FL model training.
Huang \textit{et al.}~\cite{Huang2021AnEC} introduced a long-term fairness constraint by setting a guaranteed rate for each FL client to be selected.
However, the notion of fairness in~\cite{Huang2021AnEC} is not properly constructed. Fairness in their context refers to how likely the FL server randomly selects clients in each round. 
In FL, clients do not contribute equally to the FL model due to data and resource heterogeneity. Such a strictly equal selection scheme can be perceived as unfair by clients who contribute more significantly to the FL model than others.
Such a problem, referred to as the free-rider issue in FL \cite{Lyu2019TowardsFA}, can discourage high-quality clients from participating in FL~\cite{Zhang2020HierarchicallyFF}, which is highly harmful.
Song \textit{et al.}~\cite{SongReputation22} proposed a reputation-based FL client selection scheme with fairness constraints.
The notion of fairness adopted is selection fairness~\cite{shi2021survey}, which prefers clients with higher reputation values while also providing opportunities for other clients to join FL.
However, \cite{SongReputation22} does not provide opportunities for clients to redeem their reputations as it adopts a threshold-based approach.
In reputation-based best-effort systems, it is natural for the perceived reputation of the participants to fluctuate. 
If a client's reputation falls below the threshold at any point in time, it will be excluded from all future FL tasks.

The proposed FairFedCS advances the FL client selection literature by providing a framework to balance performance and fairness considerations according to the preference of the system administrator, while enabling FL clients to redeem their reputations through self-improvement at any point in time.
\vspace*{-4mm}
\section{The Proposed Approach}
\begin{figure}[ht]
    \centering
    \includegraphics[width=1\columnwidth]{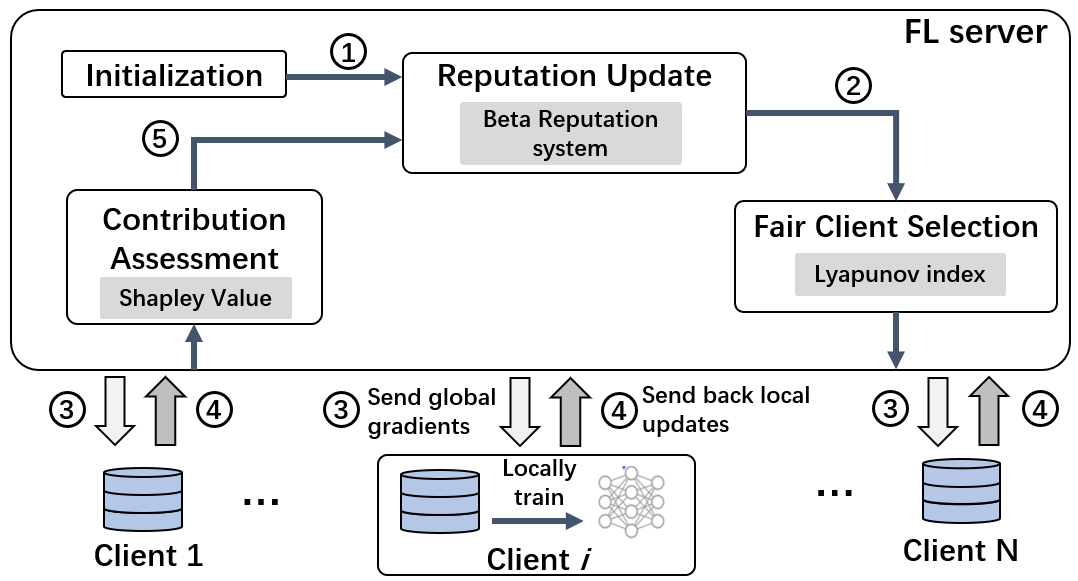}
    \caption{The FairFedCS system architecture.}
    \label{fig:FairFedCS}
\end{figure}

Under FairFedCS, part of the FL server's goal is to give preference to high-quality FL clients (in terms of data quality). However, since the local datasets are not accessible directly by the FL server, it cannot know each client's data distribution and quality.  
Hence, we design a reputation model to estimate the potential contribution by a given client, which is used as a reference to select high-quality clients while ensuring fair treatment of FL clients (as illustrated in Fig. \ref{fig:FairFedCS}).
\vspace*{-4mm}
\subsection{Contribution-based FL Client Reputation Model} \label{beta_reptataion_model}
The contribution-based FL client reputation model is designed based on the Beta Reputation System (BRS)~\cite{Jøsang02thebeta} and Shapley Value (SV)~\cite{shapley1997value}.
BRS calculates each client's reputation based on its previous performance track records.
For each training round, when the server receives the latest local updates from a client $i$ and aggregates the local updates into a global model, one of the following two scenarios could occur: 
1) the global model accuracy increases, which means client $i$ made a \textit{positive} contribution $a$ in this round; or
2) the global model accuracy decreases, which means client $i$ made a \textit{negative} contribution $b$ in this round.

The reputation model maintains a reputation table in the FL server (based on clients' contributions).
For each client $i$, the server records its number of positive contributions $a_i$ and negative contributions $b_i$. 
With this information, the server predicts the behavior of client $i$ (if it is selected) for the next round. When there is no prior information, the reputation of client $i$ is initialized to the uniform distribution $P(i)=uni(0,1)=Beta(1,1)$ by default. BRS computes the reputation value $r_i$ of client $i$ as:
\begin{equation}
    r_i = \mathbb{E}[Beta(a_i+1, b_i+1)] = \frac{a_i+1}{a_i+b_i+2}.
\end{equation}
\vspace*{-3mm}

In order to accurately estimate clients' reputations, a reliable evaluation of their contributions to the FL model is key.
Since the FL server can only access the local model updates submitted by the FL clients during each round of training, SV-based approaches that evaluate each client $i$'s impact on the global model in the given coalition are commonly adopted. 
SV, a marginal contribution-based scheme, is a solution concept in cooperative game theory. 
Let there be $n$ FL clients, 
$S \subseteq N = \{1, 2, \dots, n\}$ is a multi-set.
The FL model trained on $S$ is denoted by $w_{S}$.
The accuracy of a model $w$ evaluated on a standard test set is denoted as $f(w)$. 
The SV $\phi_i$ is used to calculate the contribution of FL client $i$ is:
\begin{equation} \label{SVequation}
\phi_i = C  {\sum_{S \subseteq N \backslash\{ i\}} \frac{f(w_{S\cup \{i\}}) - f(w_{S})}{\bigl(\begin{smallmatrix} n-1 \\ |S|  \end{smallmatrix}\bigr)} }
\end{equation}
where $C$ is a constant.

By averaging the sum of the marginal contributions over all subsets of $S$ not containing $i$,
SV reflects the contribution of client $i$ as a result of only its local data, regardless of its order of joining the FL coalition. Hence, it produces fair client contribution evaluations. However, since the computational complexity of calculating SV increases exponentially, we leverage the GTG-Shapley approach \cite{GTG-Shapley}, which provides accurate estimation of the SVs in a highly efficient manner. 
During FL model aggregation, the FL server characterizes each client's behavior based on its SV.
If $\phi_i \geqslant 0$, client $i$ has made a positive contribution, which means that the global accuracy increases after aggregating the local updates of client $i$, so we update the reputation by $a_i^{new} = a_i+1$. Otherwise, it is deemed to have made a negative contribution, and we update the reputation by $b_i^{new} = b_i+1$. 


\vspace*{-1mm}
\subsection{Balancing Fairness and Performance Goals}
The proposed FairFedCS approach leverages Lyapunov optimization to jointly consider the fair treatment of FL clients and FL model performance to trade-off between the two goals. 
Lyapunov optimization relies on a properly designed Lyapunov function to control a given dynamic system optimally~\cite{meyn2012markov}. 
The key here is to convert the dual objectives into a single objective function to be optimized without manually setting decision thresholds.

To achieve this design goal, we first introduce a virtual queue for each FL client. The queuing dynamics of a given FL client $i$'s virtual queue can be expressed as:
\begin{equation}
    Q_i(t+1) = \max[0, Q_i(t) + c_i(t) - x_i(t)]. \label{queuing_dynamics}
\end{equation}
It reflects the level of unfairness accumulating on client $i$ over time based on tracking the opportunities for $i$ to join FL training in the context of its reputation.
$x_i(t) \in \{0,1\}$ indicates whether $i$ has been selected to join FL training for the current round $t$ (1=yes, 0=no). 
$c_i(t) \in [0,1]$ is the rate at which the unfairness against $i$ is accumulated at training round $t$. 
\begin{equation}
c_i(t) = \epsilon r_i(t)1_{[x_i(t)=0]}
\end{equation}
where $\epsilon\in[0,1]$ is a discount factor to control the rate at which the unfairness is accumulated in general. In this work, we set $\epsilon = \frac{m}{N}$ where $m$ is the number of FL clients required to join FL training in each round, and $N$ is the total number of clients available.
$1_{[\text{condition}]}$ is an indicator function. Its value is 1 if and only if [condition] is true; otherwise, it evaluates to 0. The dynamics of $Q_i(t)$ are as follows:
\begin{itemize}[topsep=1pt,itemsep=1pt,partopsep=1pt, parsep=1pt]
    \item $Q_i(t)$ increases in such way that if $x_i(t)=0$ at the time when the value of $Q_i(t)$ is updated, it grows by $\epsilon r_i(t)$. This ensures that $Q_i(t)$ keeps increasing if $i$ is not selected to join FL training.
    \item $Q_i(t)$ is reduced by the value of $x_i(t)$ at any $t$.
\end{itemize}
Intuitively, for a client $i$ with a higher reputation, its $Q_i(t)$ grows at a higher rate if it is not selected to join FL training.

One objective of FairFedCS is to help the server optimize the client selection choice $\boldsymbol{X_t}=\{x_1(t), x_2(t), \cdots, x_n(t)\}$ for each training round so that $\{Q_i(t)|\forall i, \forall t\}$ are stable (i.e., not grow to infinity). 
To prevent the queues from growing towards infinity, all virtual queues must remain mean rate stable across the FL process (i.e., $\lim_{T \to \infty} \mathbb{E}[Q_i(T)]/T = 0$)~\cite{queueingNeely}. Thus, the growth rate of all $Q_i(t)$ shall not be positive:
\begin{equation}
    \lim_{T \to \infty} \frac{1}{T}{\sum_{t=1}^T} \mathbb{E} [c_i(t) - x_i(t)] \leqslant 0.\label{long_term_constraint}
\end{equation}
Eq. \eqref{long_term_constraint} is a ``soft'' constraint that allows short-term violations. 

We then leverage Lyapunov optimization~\cite{queueingNeely} to bound every increase of $Q_i(t)$ to satisfy Eq. \eqref{long_term_constraint} in a best-effort manner. We first define a quadratic Lyapunov function, $L(\Theta(t))$:
\begin{equation}
    L(\Theta(t)) = \frac{1}{2} {\sum_{i=1}^N} Q_i^2(t) \geqslant 0.
\end{equation}
It is used to measure the distribution of unfairness among FL clients. A small value of $L(\Theta(t))$ is desirable as it indicates that the general level of unfairness is low and evenly distributed among clients (in view of their reputation).

Then, we formulate the \textit{Lyapunov drift}, $\Delta(\Theta(t))$, to measure the expected increase of $L(\Theta(t)$ in one time step:
\begin{equation}
    \Delta(\Theta(t)) = \mathbb{E}[L(\Theta(t+1))-L(\Theta(t))|\Theta(t)].\label{LD1} 
\end{equation}
By minimizing $\Delta(\Theta(t))$, we aim to limit the growth of the all $Q_i(t)$ by dynamically selecting clients to join FL training.
$\Delta(\Theta(t))$ can be further derived into
\begin{equation}
    \Delta(\Theta(t))\leqslant {\sum_{i=1}^N} \left(Q_t(t)[c_i(t) - x_i(t)] + {\theta} \right)\label{surprise}
\end{equation}
where $\theta = \frac{1}{2}(c_{i}^{\max})^2 + \frac{1}{2}(x_{i}^{\max})^2$. $c_i^{\max}=1$ and $x_i^{\max}=1$ are the upper limits for $c_i(t)$ and $x_i(t)$ for all $i$ and $t$. 

Another objective of FairFedCS is to ensure that the final FL model achieves good performance. Selecting clients with higher reputation values (usually reflecting high data quality) is more likely to produce a global FL model with better performance. To this end, we define $U(t)$ to represent the expected overall utility of a strategy (i.e., the sum of the reputation values of the selected FL clients):  
\begin{equation}
    U(t) = {\sum_{i=1}^N} r_i(t) x_i(t).\label{utility}
\end{equation}

To jointly optimize the dual objectives of fairness and performance, we design a (\textit{utility - drift}) objective function:
\begin{equation} \label{utility-minus-drift}
    \begin{aligned}
        \sigma U(t) - \Delta(\Theta(t))
    \end{aligned}
\end{equation}
where $\sigma>0$ is a control parameter that enables system administrators to indicate their relative preferences between these two objectives. 
By substituting Eq. \eqref{surprise} and Eq. \eqref{utility} into Eq. \eqref{utility-minus-drift}, we have:
\begin{equation}
    \begin{aligned}
    &\frac{1}{T}{\sum_{t=0}^{T-1}}\left(\sigma \mathbb{E}[U(t)|\Theta(t)] - \Delta (\Theta(t)) \right) \geqslant \\
    &\frac{1}{T} {\sum_{t=0}^{T-1}} {\sum_{i=1}^{N}} \left(\sigma r_i(t) x_i(t) - Q_i(t) c_i(t) + Q_i(t) x_i(t) - {\theta} \right).
    \end{aligned}
\end{equation}
As FairFedCS aims to optimize the client selection choices $\boldsymbol{X_t}=\{x_1(t), x_2(t), \cdots, x_n(t)\}$ for each training round, 
we focus on the terms containing the control variable $x_i(t)$. Thus, the objective function can be simplified as:
\begin{equation}\label{ob_fn}
    \begin{aligned}
     \max_{\boldsymbol{X_t}} \quad &\frac{1}{T}{\sum_{t=0}^{T-1}} {\sum_{i=1}^{N}} x_i(t)[\sigma r_i(t) + Q_i(t)],\\
     s.t. \quad & x_i(t) \in \{0,1\}. \\
    \end{aligned}
\end{equation}

To solve Eq. \eqref{ob_fn}, we first propose the \textit{Client Suitability Index} (CSI), $\Psi_i(t)$, as:
\begin{equation}
\Psi_i(t) = \sigma r_i(t) + Q_i(t).
\end{equation}
With this index, FairFedCS ranks FL clients in descending order of their CSI values to facilitate client selection.
For each round, the FL server selects the top-$m$ clients with the highest CSI values for model training, and sends the global model parameters to them.
Then, the FL server updates the virtual queues for all the clients. 


Compared to other reputation-based FL client selection strategies, FairFedCS is more advantageous as it does not rely on a reputation threshold to qualify clients' eligibility to be selected for FL training. 
In this way, it does not completely exclude FL clients with perceived low reputations, giving them chances to contribute to FL and redeem their reputation, thereby further enhancing fairness. In addition, FL clients with hard samples might be mistakenly given a low reputation during the initial stages of training~\cite{HardSampleNoise}. FairFedCS does not preclude an FL system from leveraging their contributions at later stages of training (in which hard samples can further boost model performance).

\begin{table*}[ht]
\centering
\caption{Performance comparison (in terms of test accuracy) under Scenario 1 (S1) and Scenario 2 (S2).}
\label{table:fair_acc}
\resizebox*{1\linewidth}{!}{
\begin{tabular}{|l|cccccc|cccccc|}
\hline
\multirow{3}{*}{Method} & \multicolumn{6}{c|}{fashion-MNIST}                                                                                                                                                                           & \multicolumn{6}{c|}{CIFAR-10}                                                                                                                                                                                 \\ \cline{2-13} 
                        & \multicolumn{2}{c|}{S1}                                                   & \multicolumn{2}{c|}{S2 (p=10\%)}                                        & \multicolumn{2}{c|}{S2 (p=20\%)}                   & \multicolumn{2}{c|}{S1}                                                   & \multicolumn{2}{c|}{S2 (p=10\%)}                                        & \multicolumn{2}{c|}{S2   (p=20\%)}                  \\ \cline{2-13} 
                        & \multicolumn{1}{c|}{JFI}            & \multicolumn{1}{c|}{Acc (\%)}            & \multicolumn{1}{c|}{JFI}            & \multicolumn{1}{c|}{Acc (\%)}            & \multicolumn{1}{c|}{JFI}            & Acc (\%)            & \multicolumn{1}{c|}{JFI}            & \multicolumn{1}{c|}{Acc (\%)}            & \multicolumn{1}{c|}{JFI}            & \multicolumn{1}{c|}{Acc (\%)}            & \multicolumn{1}{c|}{JFI}            & Acc (\%)            \\ \hline
Random                  & \multicolumn{1}{c|}{0.556}          & \multicolumn{1}{c|}{86.63}          & \multicolumn{1}{c|}{0.543}          & \multicolumn{1}{c|}{80.52}          & \multicolumn{1}{c|}{0.540}          & 80.81          & \multicolumn{1}{c|}{0.545}          & \multicolumn{1}{c|}{53.95}          & \multicolumn{1}{c|}{0.523}          & \multicolumn{1}{c|}{51.34}          & \multicolumn{1}{c|}{0.532}          & 50.31          \\
Greedy                  & \multicolumn{1}{c|}{0.236}          & \multicolumn{1}{c|}{85.18}          & \multicolumn{1}{c|}{0.266}          & \multicolumn{1}{c|}{81.54}          & \multicolumn{1}{c|}{0.210}          & 78.50          & \multicolumn{1}{c|}{0.205}          & \multicolumn{1}{c|}{47.66}          & \multicolumn{1}{c|}{0.187}          & \multicolumn{1}{c|}{44.37}          & \multicolumn{1}{c|}{0.170}          & 41.78          \\
RBCS-F                  & \multicolumn{1}{c|}{0.559}          & \multicolumn{1}{c|}{85.65}          & \multicolumn{1}{c|}{0.550}          & \multicolumn{1}{c|}{80.10}          & \multicolumn{1}{c|}{0.510}          & 79.52          & \multicolumn{1}{c|}{0.566}          & \multicolumn{1}{c|}{52.03}          & \multicolumn{1}{c|}{0.527}          & \multicolumn{1}{c|}{49.55}          & \multicolumn{1}{c|}{0.544}          & 48.69          \\
RBFF                    & \multicolumn{1}{c|}{0.436}          & \multicolumn{1}{c|}{86.55}          & \multicolumn{1}{c|}{0.515}          & \multicolumn{1}{c|}{\textbf{83.19}} & \multicolumn{1}{c|}{0.499}          & 80.40          & \multicolumn{1}{c|}{0.523}          & \multicolumn{1}{c|}{55.81}          & \multicolumn{1}{c|}{0.545}          & \multicolumn{1}{c|}{53.54}          & \multicolumn{1}{c|}{0.384}          & 53.38          \\ \hline
FedFairCS               & \multicolumn{1}{c|}{\textbf{0.776}} & \multicolumn{1}{c|}{\textbf{87.39}} & \multicolumn{1}{c|}{\textbf{0.684}} & \multicolumn{1}{c|}{83.16}          & \multicolumn{1}{c|}{\textbf{0.722}} & \textbf{83.27} & \multicolumn{1}{c|}{\textbf{0.764}} & \multicolumn{1}{c|}{56.15}          & \multicolumn{1}{c|}{\textbf{0.739}} & \multicolumn{1}{c|}{53.85}          & \multicolumn{1}{c|}{\textbf{0.749}} & 53.43          \\ 
Ablation Study          & \multicolumn{1}{c|}{0.590}          & \multicolumn{1}{c|}{87.22}          & \multicolumn{1}{c|}{0.577}          & \multicolumn{1}{c|}{81.12}          & \multicolumn{1}{c|}{0.587}          & 80.73          & \multicolumn{1}{c|}{0.591}          & \multicolumn{1}{c|}{\textbf{56.38}} & \multicolumn{1}{c|}{0.581}          & \multicolumn{1}{c|}{\textbf{54.10}} & \multicolumn{1}{c|}{0.590}          & \textbf{53.52} \\\hline
\end{tabular}
}
\end{table*}
\vspace*{-1mm}
\section{Experimental Evaluation}
\vspace*{-1mm}
\subsection{Experiment Settings}
We conduct experiments on two public multimedia datasets: fashion-MNIST and CIFAR-10. For both datasets, the training set was distributed to $N=40$ clients. Each client owns 1,100 images. 
For each round, we select $m=4$ clients from the population. We investigated two experimental scenarios:
\begin{enumerate} [topsep=1pt,itemsep=1pt,partopsep=1pt, parsep=1pt]
    \item \textbf{IID data, different noisy data percentages:} Images are randomly sampled from the training set and distributed to clients. To simulate different data quality levels, we generate different numbers of noisy data for the clients: clients $\{1, 2, 3, \cdots, 10, 11, \cdots, 20\}$ own $\{0\%, 5\%, 10\%,  \cdots, 45\%, 0\%, \cdots, 45\%\}$ noisy data.

    \item \textbf{Non-IID data, different noisy data percentages, minority clients of different data qualities:} We adjust the number of classes assigned to each client to simulate data heterogeneity. We select one class label as the minority class (e.g., digit 1). Three clients are designated as minority clients. They own images from all 10 classes, while all other clients only own images from 9 classes (except the minority class). This is to simulate the existence of hard samples. We also simulate conditions where the minority clients contain $p \in \{10\%, 20\%\}$ of noisy data. The percentages of noisy data for the remaining clients are generated following the settings in Scenario 1.
\end{enumerate}
\vspace*{-2mm}
\subsection{Comparison Approaches}
We compare FairFedCS against 4 existing approaches:
\begin{enumerate}[topsep=1pt,itemsep=1pt,partopsep=1pt, parsep=1pt]
    \item \textbf{Random}: In each round, the server randomly selects $m$ clients. 

    \item \textbf{Greedy}: The server selects the top $m$ clients with the highest reputation in each round.

    \item \textbf{RBFF}: The Reputation-based FL Framework (RBFF)~\cite{SongReputation22} introduces a reputation-based scheduling policy with fairness constraints. It trades off between reputation and clients' participation rates\footnote{Participation Rate of Client $i$ = $ \frac{\text{Number of participation by Client }i}{\text{Total number of participation by all clients}}$}. 

    \item \textbf{RBCS-F}: Reputation-based Client Selection with Fairness (RBCS-F)~\cite{Huang2021AnEC} jointly considers training efficiency and fairness. It is also based on Lyapunov optimization. Differing from FairFedCS, RBCS-F focuses on the training efficiency of clients in terms of the model exchange time. Moreover, the fairness constraint of RBCS-F is to set a constant selection rate, $\eta\in[0,1]$, for every client. Hence, the virtual queues in RBCS-F are updated by:
    \begin{equation}\label{huang_queue}
        Q_i(t+1) = \max[0, Q_i(t) + \eta - x_i(t)].
    \end{equation}
    We set $\eta = \frac{m}{N}$,  same as $\epsilon$ setting in FairFedCS.
\end{enumerate}

We also design an ablated version of FairFedCS by replacing Eq. \eqref{queuing_dynamics} with that of RBCS-F (Eq. \eqref{huang_queue}). This is to investigate the effect of incorporating reputation into the virtual queues on client selection fairness.

We set up an FL environment with \textit{PyTorch}. We use two convolutional neural network (CNN) models from \cite{Huang2021AnEC} to perform the classification tasks on fashion-MNIST and CIFAR-10.
We set the learning rate to 0.01 for fashion-MNIST and 0.005 for CIFAR-10. The control parameter $\sigma$ is set to be 0.6.

\vspace*{-2mm}
\subsection{Evaluation Metrics}
We evaluate FedFairCS based on the following metrics:
\begin{enumerate}[topsep=1pt,itemsep=1pt,partopsep=1pt, parsep=1pt]
    \item \textbf{Test Accuracy at Convergence (Acc)}: 
    In order to prevent overfitting, we perform early stopping in the experiments when the test loss does not improve for 20 rounds. The FL server then evaluates the FL model on its test set to calculate the test accuracy. 
    
    \item \textbf{Selection Fairness based on Jain's Fairness Index (JFI)}: We adopt Jain's Fairness Index (JFI) \cite{Jain1998AQM, cho2020bandit, shi2021survey} to evaluate the degree of selection fairness achieved by each approach after model convergence to investigate the level of fairness. It is formulated as follows:
    \begin{equation}
        \text{JFI} = \frac{({\sum_{i=1}^n \frac{y_i}{z_i}})^2}{n \times \sum_{i=1}^n (\frac{y_i}{z_i})^2}
    \end{equation}
    where 
    $y_i$ represents the total number of participation times by client $i$ by the end of the training, and $z_i$ represents the ground truth data quality of client $i$.
    For the evaluation method for client quality, we follow the protocol in~\cite{deng2021auction}.
    The data quality of client $i$ is computed based on the number of image classes $n\_class_i$ and the percentage of noisy data $p\_noisy_i$ (i.e., $z_i = n\_class_i \times (1 - p\_noisy_i$)). We then use min-max scaling to normalize the range of $z_i$ from $[0.5, 1]$ to $[0, 1]$, such that the notion of selection fairness can be applied to different datasets under different noisy data settings.
    JFI ranges from 0 (the most unfair) to 1 (the most fair).
\end{enumerate}

\vspace*{-2mm}
\subsection{Results and Discussion}
We analyze the performance of the comparison approaches in terms of fairness and test accuracy. 
The averaged results after twenty times of experiments are shown in Table \ref{table:fair_acc}.

In terms of fairness, it can be observed that 
FairFedCS significantly outperforms all the baselines.
It achieved 19.6\% higher JRI on average than the best-performing baseline, RBCS-F.
This implies that FairFedCS has brought significant improvement in selection fairness. Greedy performs the worst as it focuses on selecting only a few high-quality clients without giving opportunities to the rest.
FairFedCS has also achieved higher fairness than its ablated version, demonstrating that our key design of incorporating reputation into the virtual queues enhances selection fairness. 

Intuitively, improving selection fairness is likely to result in drops in test accuracy as it increases the probability of low-quality clients being selected, which negatively impacts FL model performance. However, the experiment results show that FairFedCS maintains a satisfactory test accuracy while achieving significant gains in fairness. Table \ref{table:fair_acc} shows that FairFedCS consistently achieves top-2 ranks in test accuracy under all conditions. On average, it achieves 0.73\% higher test accuracy than the best-performing baseline, RBFF.


We can also observe that FairFedCs beats all baselines in test accuracy in both datasets when $p = 20\%$.
It outperforms the best-performing baseline approach by 2.71\% over fashion-MNIST. This demonstrates the key advantages of FairFedCS: 1) It is capable of identifying minority clients with relatively low-quality data; and 2) it selects minority clients with reasonable frequencies such that the final FL model is sufficiently exposed to the minority class while not significantly affected by noisy data.

\vspace*{-1mm}
\section{Conclusions and Future Work}
This paper proposed FairFedCS, a dynamic FL client selection approach to balance model performance with selection fairness.
Under FairFedCS, no client is excluded from being selected, and their selection probabilities depend on their potential contributions. 
Extensive experiments on real-world datasets demonstrate that FairFedCS significantly enhances selection fairness while maintaining a high level of test accuracy. 
It achieves 19.6\% higher fairness and 0.73\% higher test accuracy on average than the best-performing baseline.

In subsequent research, we will further adapt FairFedCS to dynamic FL marketplaces with multiple FL servers competing for FL clients to investigate how to ensure selection fairness under non-monopoly settings.

\section{Acknowledgments}
This research/project is supported by the National Research Foundation, Singapore and DSO National Laboratories under the AI Singapore Programme (AISG Award No: AISG2-RP-2020-019); Alibaba Group through Alibaba Innovative Research (AIR) Program and Alibaba-NTU Singapore Joint Research Institute (JRI), Nanyang Technological University, Singapore; the RIE 2020 Advanced Manufacturing and Engineering (AME) Programmatic Fund (No. A20G8b0102), Singapore; Nanyang Assistant Professorship (NAP); and Future Communications Research \& Development Programme (FCP-NTU-RG-2021-014).
\small
\bibliographystyle{IEEEbib}
\bibliography{icme2021template}

\end{document}